\documentclass[letterpaper]{article} 
\usepackage{aaai2026}  
\usepackage{times}  
\usepackage{helvet}  
\usepackage{courier}  
\usepackage[hyphens]{url}  
\usepackage{graphicx} 
\usepackage{natbib}  
\usepackage{caption} 
\usepackage{algorithm}

\usepackage{listings}
\DeclareCaptionStyle{ruled}{labelfont=normalfont,labelsep=colon,strut=off} 
\floatstyle{ruled}
\newfloat{listing}{tb}{lst}{}
\floatname{listing}{Listing}

\usepackage{multirow}
\usepackage{amsmath}
\usepackage{url}
\usepackage{soul}
\usepackage{multirow}
\usepackage{algpseudocode}
\usepackage{amsmath}
\usepackage{amssymb}
\usepackage{booktabs}
\usepackage[percent]{overpic}
\usepackage{makecell}

\usepackage{enumitem}

\usepackage[capitalize]{cleveref}
\crefname{section}{Sec.}{Secs.}
\Crefname{section}{Section}{Sections}
\Crefname{table}{Table}{Tables}
\crefname{table}{Tab.}{Tabs.}

\nocopyright 

\makeatletter
\def\@fnsymbol#1{\ensuremath{\ifcase#1\or \dagger\or \ddagger\or
\mathsection\or \mathparagraph\or \|\or **\or \dagger\dagger
\or \ddagger\ddagger \else\@ctrerr\fi}}
\makeatother

\title{SOSControl: Enhancing Human Motion Generation through Saliency-Aware Symbolic Orientation and Timing Control}

\author{
    Ho Yin Au,
    Junkun Jiang,
    Jie Chen\thanks{Corresponding Author}
}
\affiliations{
    Department of Computer Science, Hong Kong Baptist University\\
    \{cshyau, csjkjiang, chenjie\}@comp.hkbu.edu.hk
}

\begin{document}

\maketitle

\begin{abstract}
Traditional text-to-motion frameworks often lack precise control, and existing approaches based on joint keyframe locations provide only positional guidance, making it challenging and unintuitive to specify body part orientations and motion timing. To address these limitations, we introduce the Salient Orientation Symbolic (SOS) script, a programmable symbolic framework for specifying body part orientations and motion timing at keyframes.
We further propose an automatic SOS extraction pipeline that employs temporally-constrained agglomerative clustering for frame saliency detection and a Saliency-based Masking Scheme (SMS) to generate sparse, interpretable SOS scripts directly from motion data. Moreover, we present the SOSControl framework, which treats the available orientation symbols in the sparse SOS script as salient and prioritizes satisfying these constraints during motion generation. By incorporating SMS-based data augmentation and gradient-based iterative optimization, the framework enhances alignment with user-specified constraints. Additionally, it employs a ControlNet-based ACTOR-PAE Decoder to ensure smooth and natural motion outputs.
Extensive experiments demonstrate that the SOS extraction pipeline generates human-interpretable scripts with symbolic annotations at salient keyframes, while the SOSControl framework outperforms existing baselines in motion quality, controllability, and generalizability with respect to motion timing and body part orientation control. 
\end{abstract}

\begin{links}
    \link{Code}{https://github.com/asdryau/SOSControl}
\end{links}

\section{Introduction}

Text-conditioned human motion generation has received substantial research focus due to its potential to produce diverse humanoid motions guided by text prompts, with promising applications in media content creation, robotics, and human-AI collaboration. However, since text descriptions are often subjective and ambiguous, traditional text-to-motion frameworks lack precise control over body part orientation and timing, prompting the integration of additional conditioning signals to enhance motion controllability.

Recent research has explored the use of joint keyframe locations to enhance controllability in motion generation. However, this approach often offers limited control over body part orientation and timing, and accurately defining plausible keyframe locations remains complex. For example, specifying only the final fist location for a \textit{squatted forward punch} may produce incorrect arm orientation, such as an unintended \textit{lower punch}, if the system adjusts primarily through shoulder rotation rather than coordinating the entire body. Also, the model may misinterpret end locations as intermediate waypoints, resulting in overshooting and disrupting the intended punch timing. Moreover, ensuring that specified 3D joint locations are accurately placed and physically executable requires extensive manual adjustments in animation tools, such as frequent switching between camera views, and a thorough understanding of motion dynamics (e.g., adjusting for appropriate movement speed and physical balance), making the workflow time-consuming and impractical for industrial animation pipelines.

\begin{figure}[t]
    \centering
    \includegraphics[width=1.0\linewidth]{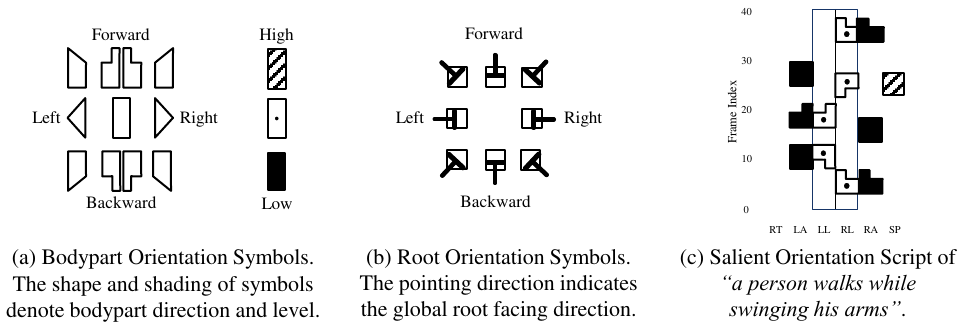}
    \caption{Salient Orientation Script (SOS) Illustration: (a) Body Part and (b) Root Orientation Symbols specify keyframe states for effective motion control. (c) SOS example as a staff highlights its programmable interface potential.}
    \label{fig:symbol_illust}
\end{figure}

To tackle these challenges, we introduce the Salient Orientation Symbolic (SOS) script, a programmable symbolic framework designed to define and represent body part orientations and motion timing within motion sequences. Inspired by Labanotation~\cite{guest2013labanotation}, the SOS script uses orientation symbols to annotate individual body parts at high saliency keyframes. As shown in Fig.~\ref{fig:symbol_illust}, the SOS script is represented as a symbolic staff, offering an intuitive, programmable interface for adjusting orientation and timing through drag-and-drop symbol placement. Additionally, we propose an automatic extraction pipeline that generates SOS scripts directly from motion sequences, utilizing temporally-constrained agglomerative clustering to extract frame saliency. Leveraging this saliency information, our Saliency-based Masking Scheme (SMS) adaptively filters the extracted orientation features below user-defined saliency thresholds, highlighting key motion moments and producing a sparse, human-interpretable SOS script. Users can adjust these thresholds at test time to interactively control the SOS script’s sparsity for customizable visualization.

Building on this symbolic representation, we introduce SOSControl, which integrates the SOS script with language-guided motion generation. 
Since SOS scripts emphasize retaining high-saliency keyframes, the SOSControl framework considers the available orientation symbols in the input sparse SOS script as salient and prioritizes satisfying these constraints during the motion generation process.
Extending ControlNet-based motion diffusion methods, SOSControl incorporates SMS into data preprocessing and augmentation, enabling the model to prioritize retained symbols and precisely synchronize motion peaks with intended timings. The differentiable orientation feature extraction pipeline further supports gradient-based iterative optimization, ensuring precise alignment with input orientations.
Additionally, the ControlNet-based ACTOR-PAE Decoder regularizes motion outputs for smoothness and naturalness, allowing stable iterative optimization during inference.
To our knowledge, this is the first approach to use saliency information from agglomerative clustering for enhanced motion control. Our contributions are as follows:

\begin{itemize}
    \item We introduce the SOS script, a programmable symbolic framework with an intuitive staff-based interface for representing body part orientations and motion timing using orientation symbols annotated at keyframes.
    \item  We propose an automatic SOS extraction pipeline, which uses temporally-constrained agglomerative clustering to identify frame saliency, and applies a Saliency-based Masking Scheme (SMS) to adaptively filter orientation features, producing sparse and interpretable SOS scripts.
    \item We present the SOSControl framework, which integrates SOS scripts into motion generation using SMS-based data augmentation, gradient-based optimization, and employs a ControlNet-based ACTOR-PAE Decoder for smooth and natural motion outputs.
\end{itemize}

\section{Related work}

\textbf{Motion Diffusion with Textual Descriptions.}
Diffusion frameworks~\cite{song2020DDIM, ho2020DDPM} have proven effective in generating high-quality outputs across diverse domains. In text-to-motion generation, foundational diffusion frameworks like MLD~\cite{chen2023mld}, MDM~\cite{tevet2023mdm}, and MotionDiffuse~\cite{zhang2024motiondiffuse} offer great extensibility by enabling precise and flexible control through detailed text descriptions analysis. AttT2M~\cite{zhong2023attt2m}, FineMoGen~\cite{zhang2023finemogen}, and CoMo~\cite{huang2024como} further enhance motion control by establishing more accurate associations between text inputs and specific body parts.
Meanwhile, GraphMotion~\cite{jin2023graphmotion} and Fg-T2M~\cite{wang2023fgt2m} utilize text-based semantic graphs to analyze input text comprehensively, uncovering detailed relationships between body parts and motion specifications to improve the accuracy and contextual relevance of the produced motion.
However, achieving detailed motion control using freeform text remains challenging, as users must verbosely specify all body part conditions, joint locations, and motion semantics in a paragraph. Moreover, the text processing pipelines in these models may filter or misinterpret instructions, resulting in generated motions that may not fully reflect the user’s intent.

\noindent\textbf{Integrating Control Signals in Motion Diffusion.}
In addition to detailed text analysis, motion diffusion frameworks can be extended to incorporate additional control signals, such as joint keyframe locations, into the generation process. For instance, PriorMDM~\cite{shafir2024priorMDM} extends MDM by iteratively imputing user-specified root trajectories during motion sampling for trajectory-guided generation. Similarly, GMD~\cite{karunratanakul2023gmd} refines diffused motion by inferring the root trajectory from the generated motion, comparing it to the user-specified trajectory, and applying gradient-based optimization for refinement.
Recent methods further enhance keyframe alignment through iterative optimization. For example, OmniControl~\cite{xie2024omnicontrol} performs diffusion-time optimization by computing gradients at each diffusion step to improve alignment with keyframe specifications, while leveraging the ControlNet architecture to ensure motion coherence among unspecified frames and joints. In contrast, TLControl~\cite{wan2024tlcontrol} trains a transformer to denoise VQ-VAE encoded body part tokens using both text and trajectory inputs, and employs test-time iterative optimization to better align the output motion with user-specified conditions.

\begin{figure*}[t]
    \centering
    \includegraphics[width=0.88\linewidth]{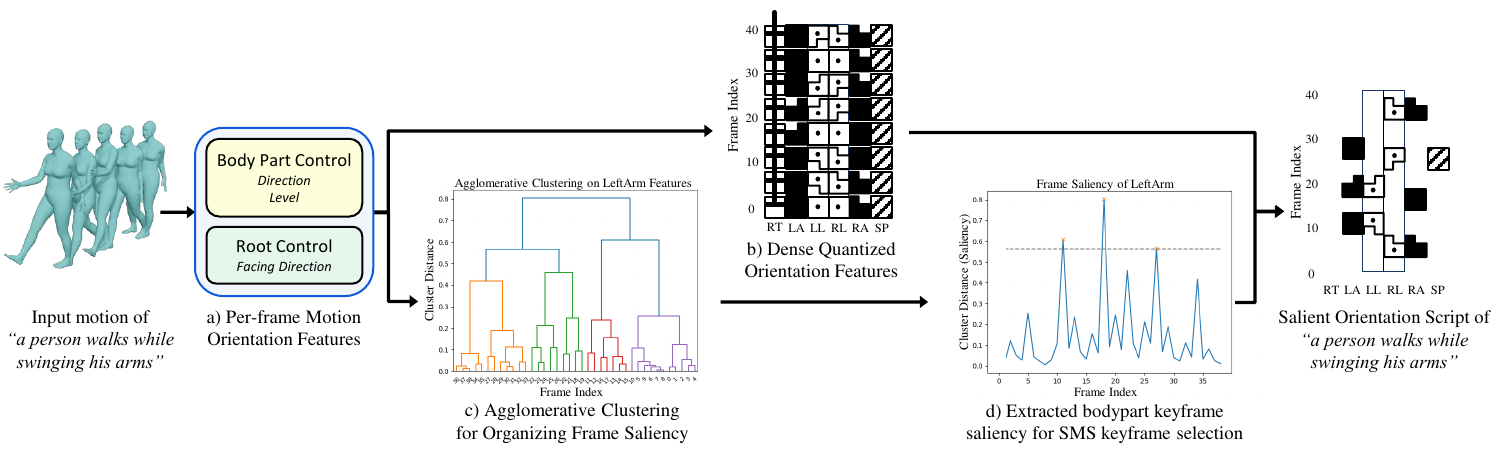}
    \caption{Overview of the SOS extraction pipeline: (a) extract per-frame orientation features, then (b) quantize them according to symbol categories from Fig.~\ref{fig:symbol_illust}. Agglomerative clustering is applied in (c) to derive the per-frame orientation feature into frame saliency shown in (d). Symbols selected from frames with saliency above a threshold compose the final SOS.}
    \label{fig:MCS_extraction}
\end{figure*}

\noindent\textbf{Abstract and Interpretable Motion Descriptors.}
Labanotation~\cite{guest2013labanotation} is a widely adopted symbolic system for recording body part movements using orientation symbols on a staff.
While direct extraction of Labanotation from motion has not yet been achieved, various studies have focused on extracting related spatial features such as joint positions and orientations. For example, PoseScript~\cite{delmas2022posescript} and CoMo~\cite{huang2024como} convert these features into body part-level text descriptions to improve text-motion alignment. Meanwhile, KP~\cite{liu2024kp}, HL~\cite{li2024handlabanotation}, and PL~\cite{jiang2024laban} transform these features into abstract, interpretable representations for motion understanding, which can also be used as control signals. However, these approaches do not address motion saliency detection, and their representations are often described in a verbose, frame-by-frame manner, making the process of programming these features highly labor-intensive.

\section{Preliminary}
\subsection{Motion Data Representation}
Due to the increasing popularity of using the SMPL humanoid model~\cite{loper2015SMPL}, human motions are typically standardized to the SMPL skeleton and represented as frame sequences comprising a global root trajectory and 24 joint rotations.
Traditional motion generative models convert SMPL motion data into 263 parameters based on the HumanML3D~\cite{guo2022t2m} pose format, which includes root-centric velocity, joint velocities, joint rotations, and foot contact information. To incorporate the missing root orientation, we add a 6D rotation~\cite{zhou2019rot6d}, resulting in a motion representation of $\mathbf{x} \in \mathbf{R}^{T\times269}$.

\subsection{Kinematic Feature Extraction}
Feature extraction algorithms for body part orientation are often developed based on the orientation concept in Labanotation. For example, KP~\cite{liu2024kp} extracts reference vectors $\mathbf{r}_t \in \mathbf{R}^{T\times3\times3}$ to define egocentric directions for each motion frame $t$, which are then used to compute the Pairwise Relative Position Phrase (PRPP)~\cite{liu2024kp}:
\begin{equation}
\mathbf{o}^{J}_t = \text{PRPP}(e^{J},a^{J})_t = (\mathbf{l}_t(e^{J}) - \mathbf{l}_t(a^{J})) \cdot \mathbf{r}_t,
\end{equation}
where $\mathbf{o}^{J}_t$ represents the relative position between the end joint $e^{J}$ and the anchor joint $a^{J}$ at frame $t$, serving as the orientation feature for body part $J$. Here, $\mathbf{l}$ denotes local joint trajectories obtained via forward kinematics with a zeroed global root trajectory. The dot product with reference vectors $\mathbf{r}_t$ transforms relative positions into egocentric directions.

\section{Salient Orientation Symbols}
Inspired by Labanotation~\cite{guest2013labanotation}, we propose the Salient Orientation Symbolic (SOS) script as an abstract motion representation designed for motion control. Using body part frame saliency extracted through temporally-constrained agglomerative clustering, SOS is depicted as a sparse and concise symbolic staff that runs vertically up. 
As shown in Fig.~\ref{fig:symbol_illust}(c), the staff contains six columns corresponding to body parts: \textit{Root} (\textit{RT}) \textit{Left Arm} (\textit{LA}), \textit{Left Leg} (\textit{LL}), \textit{Right Leg} (\textit{RL}), \textit{Right Arm} (\textit{RA}), and \textit{Spine} (\textit{SP}). 
Fig.~\ref{fig:symbol_illust}(a) presents the eight root direction symbols, while Fig.~\ref{fig:symbol_illust}(b) depicts the 26 body part orientation symbols, where shape denotes direction and shading represents level.
The vertical position of each symbol on the staff reflects the orientation state of the corresponding body part at each frame.

We developed a pipeline to extract the SOS script from motion data in four main steps: First, kinematic feature extraction transforms raw motion signals into orientation features. Second, spatial feature quantization maps these features to discrete labels. Third, hierarchical temporal saliency detection analyzes orientation features for each body part, constructing a segment tree to detect frame saliency. Finally, by applying a Saliency-based Masking Scheme (SMS) to the quantized features based on the detected saliency, the SOS script can be synthesized at various levels of granularity.

\subsection{Kinematic Feature Extraction}

We extract per-frame motion orientation features $\mathbf{o} \in \mathbf{R}^{T\times6\times3}$ to represent the orientation state in the SOS script. These features are represented as three-dimensional directional vectors for six body parts, capturing the root facing direction and the body part orientation. For the \textit{Root}, we use the horizontal facing direction $\mathbf{o}^{RT} = \mathbf{r}_t^f \in \mathbf{R}^{T\times1\times3}$, obtained by projecting the forward reference vector in $\mathbf{r}_t$ onto the ground plane. For the remaining five body parts, such as \textit{Left Arm}, the orientation $\mathbf{o}^{LA}\in \mathbf{R}^{T\times1\times3}$ is computed using the PRPP between the end joint (e.g., \textit{left wrist} $e^{LA}$) and the anchor joint (e.g., \textit{left shoulder} $a^{LA}$) for each body part.

\begin{figure*}
    \centering
    \includegraphics[width=0.88\linewidth]{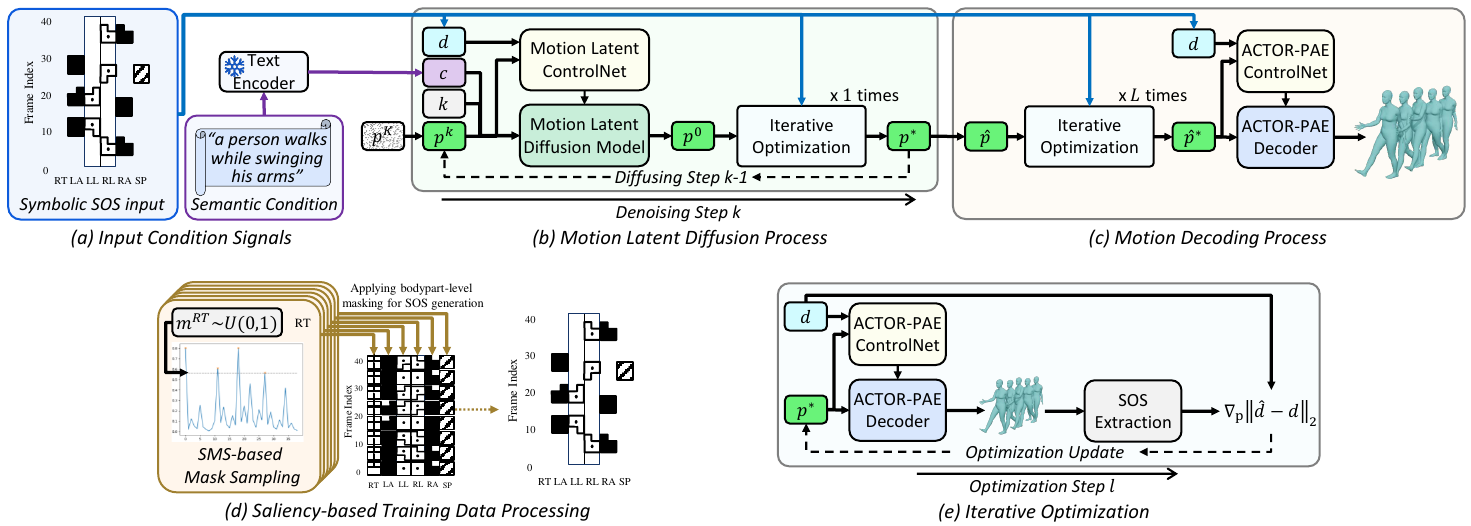}
    \caption{Overview of the SOSControl pipeline: (a) Start by obtaining a semantic condition and a Symbolic Orientation Script (SOS) from user input, (b) perform motion periodic latent diffusion, and (c) decode the resulting motion latent back to motion. Both (b) and (c) utilize ControlNet to incorporate the SOS condition into the trained models and apply iterative optimization in (e) to refine the model output, ensuring better alignment with the input SOS conditions. To improve model adaptability to diverse user-provided SOS scripts, (d) perform SMS-based mask sampling to generate SOS at varying levels of granularity, enabling the trained models to handle different saliency variations more robustly.}
    \label{fig:framework}
\end{figure*}

\subsection{Hierarchical Temporal
Saliency Detection}
Hierarchical temporal saliency detection organizes the orientation features $\mathbf{o}$ to extract body part keyframes, resulting in a sparse and easily manipulable representation.
We apply temporally‐constrained agglomerative clustering to group these features into a bottom‐up tree of connected segments. For each body part, we calculate the frame‐level central finite difference of the normalized dot product between $\mathbf{o}$ with 26 unit-norm direction vectors, using this as input to the \textit{scikit-learn} agglomerative clustering algorithm. Following the approach in Librosa~\cite{mcfee2015librosa}, we constrain the connectivity matrix of the clustering algorithm so that merging occurs only between adjacent segments, ensuring that clusters correspond to temporally contiguous motion segments. Fig.~\ref{fig:MCS_extraction}(c) shows a dendrogram of this bottom‐up segmentation, where the y-axis indicates cluster distance and the x-axis indicates frame indices.

After constructing the segmentation tree, each node comprises two temporally adjacent segment subnodes and a merging distance that indicates the value at which the segments are joined. We compute the body part keyframe saliency signal by assigning the merging distance at each node as the saliency value to the first frame of the subsequent segment subnode. Figure~\ref{fig:MCS_extraction}(d) shows the resulting saliency signal, obtained by processing all nodes in a bottom-up manner within the hierarchical tree. For instance, the saliency value at frame 18 for the \textit{Left Arm} is the highest, indicating a key moment when the arm reaches its peak during a swinging motion. Further details on saliency value extraction are provided in the Supplementary Material.

\subsection{Spatial Feature Quantization}
The spatial feature quantization process transforms $\mathbf{o}$ into discrete intervals, enabling symbolic representation as shown in Fig.~\ref{fig:MCS_extraction}(b). Based on the Labanotation orientation scheme in Fig.~\ref{fig:symbol_illust}(a), we define 26 unit-norm direction vectors $\mathbf{u}\in\mathbb{R}^{26\times3}$, each representing a specific orientations (e.g. $(1,0,0)$ for the Right-Middle symbol). To increase sensitivity to horizontal movement compared to vertical movement, we increase the weight on the upward axis (e.g. $(0,\frac{1}{\sqrt{10}},\frac{3}{\sqrt{10}})$ for the Forward-Top symbol). 
To quantize the orientation features into symbols, we compute the dot product between the orientation vector $\mathbf{o}$ and each of the 26 direction vectors in $\mathbf{u}$: 
\begin{equation}
\mathbf{q} = softmax( \frac{\mathbf{o}}{||\mathbf{o}||} \cdot \mathbf{u}^T) \cdot \mathbf{u},
\end{equation}
where the softmax operation ensures the quantization operation remains differentiable, allowing for iterative optimization during diffusion-time and test-time. For symbol recognition and visualization, the softmax can be replaced with an argmax operation. Finally, by selecting symbols from the quantized orientation features at keyframes identified by temporal saliency detection, we synthesize the final SOS staff, as shown on the right in Fig.~\ref{fig:MCS_extraction}.

\subsection{Saliency-based Masking Scheme}

Based on the body part keyframe saliency extracted from agglomerative clustering, we apply a Saliency-based Masking Scheme (SMS) that selects body part keyframes by masking all the body part frames with saliency values below a specified threshold. For instance, setting the threshold to 0.7 of the maximum saliency across all body parts results in the selection of three LeftArm keyframes, as shown in Fig.~\ref{fig:MCS_extraction}(d). This enables customizable levels of detail in the SOS script, supporting analytical visualization and data augmentation.

\section{Motion Periodic Latent Diffusion \\  with Saliency Orientation Script}

To facilitate motion generation from the SOS script, we propose a two-stage framework, as shown in Fig.~\ref{fig:framework}. In the first stage, the motion periodic latent $\mathbf{p}$ is denoised using the input text and SOS script. In the second stage, the denoised latent is decoded back into human motion. To enhance alignment between the decoded motion and the input SOS, we integrate ControlNet and iterative optimization into both motion diffusion and decoding processes. In the following sections, we first introduce the basic framework for periodic latent diffusion. We then describe the ControlNet adaptation and iterative optimization strategies used to align the generated motion with the input SOS script. Finally, we present a saliency‐based training data processing approach, which enables the model to accommodate user-provided SOS scripts with varying levels of granularity.

\subsection{Basic Periodic Latent Diffusion}
We perform motion latent diffusion in a periodic latent space defined by the ACTOR-PAE, which promotes the generation of smooth, natural motions and provides a regularized foundation for iterative optimization during motion decoding.

\subsubsection{ACTOR-PAE and Periodic Latent Space.}
Following the approach in \cite{au2025tpdm}, we construct the ACTOR-PAE model by integrating the periodic parameterization from PAE~\cite{starke2022deepphase} with the motion autoencoder architecture from ACTOR~\cite{petrovich2021actor}. Specifically, the ACTOR-PAE encoder $\mathcal{P}_E$ processes motion $\mathbf{x}$ into four phase parameters $\mathbf{f},\mathbf{a},\mathbf{b},\mathbf{s} \in \mathbb{R}^{P}$.
Following PAE~\cite{starke2022deepphase}, these parameters are used to generate a periodic signal $\mathbf{p} \in \mathbb{R}^{T \times P}$ as follows:
\begin{equation}\label{eq:phase_reparam}
\mathbf{p} = \mathbf{a} \sin (\mathbf{f} \cdot (N-\mathbf{s}))+\mathbf{b},
\end{equation}
where $N \in \mathbb{R}^{T}$ denotes the time difference of each frame relative to the center of the motion sequence. 
The ACTOR-PAE decoder $\mathcal{P}_D$ then utilizes $\mathbf{p}$ to reconstruct the motion $\mathbf{\hat{x}}$. The model is trained using mean squared error (MSE) loss.

\subsubsection{Periodic Latent Diffusion.}
Following MDM~\cite{tevet2023mdm}, we develop the Motion Latent Diffusion Model $\mathcal{D}^-$, which adopts a transformer encoder architecture to denoise the periodic latent $\mathbf{p}^k$ at diffusion step $k$, conditioned on the input text $\mathbf{c}$.
The training losses for $\mathcal{D}^-$ is as follows:
\begin{equation}
\mathcal{L}_{\mathcal{D}^-} = ||\mathbf{p}^0 - \mathcal{D}^-(k, \mathbf{c}, \mathbf{p}^k)||_2.
\end{equation}
During inference, the model predicts the clean latent $\mathbf{p}^0$ from the diffused latent $\mathbf{p}^k$ at each diffusion step $k$. DDIMScheduler~\cite{song2020DDIM} is employed to diffuse $\mathbf{p}^0$ back to $\mathbf{p}^{k-1}$ for the next diffusion step. After $K$ diffusion iterations, the final clean latent $\mathbf{p}^0$ is decoded into motion $\mathbf{\hat{x}}$ using the ACTOR-PAE decoder $\mathcal{P}_D$. 

\subsection{Injecting SOS Guidance to the Diffusion Process.}
In the basic periodic latent diffusion framework, $\mathcal{D}^-$ serves only as a foundational model and does not condition on the SOS input $\mathbf{d}$. To address this, we incorporate ControlNet and iterative optimization during both diffusion-time (Fig.~\ref{fig:framework}(b)) and test-time (Fig.~\ref{fig:framework}(c)), enabling the motion generation process to be guided and aligned with the input SOS scripts.

\subsubsection{ControlNet Adaptation to the Diffusion Model.}
After training $\mathcal{D}^-$, we adapt the model to incorporate the SOS input $\mathbf{d}$ following the approach described in ControlNet~\cite{zhang2023controlnet} and OmniControl~\cite{xie2024omnicontrol}. This yields $\mathcal{D}^+$, which consists of both the frozen original Motion Latent Diffusion Model $\mathcal{D}^-$ and a trainable copy of $\mathcal{D}^-$, referred to as the Motion Latent ControlNet. The trainable component is then finetuned to convert $\mathbf{d}$ into a guidance signal for the diffusion process. The training losses for $\mathcal{D}^+$ is as follows:
\begin{equation}
\mathcal{L}_{\mathcal{D}^+} = ||\mathbf{p}^0 - \mathcal{D}^+(k, \mathbf{c}, \mathbf{d}, \mathbf{p}^k)||_2.
\end{equation}
In addition to the Diffusion Model $\mathcal{D}^-$, ACTOR-PAE Decoder $\mathcal{P}_D$ also employs the transformer encoder architecture, enabling the application of the same ControlNet adaptation to obtain $\mathcal{P}_D^+$. By leveraging the guidance signal derived from the input SOS script $\mathbf{d}$ during motion decoding, the decoded motion becomes better aligned with the input. This enhanced alignment also leads to improved performance in the iterative optimization process.

\subsubsection{Iterative Optimization}
The denoised periodic signal $\mathbf{p}$ is decoded by $\mathcal{P}_D^+$ to produce the output motion $\mathbf{\hat{x}} = \mathcal{P}_D^+(\mathbf{p}^*, \mathbf{d}))$. The quantized orientation features $\mathbf{\hat{q}}$ are then estimated from $\mathbf{\hat{x}}$ using the extraction pipeline illustrated in Fig.~\ref{fig:MCS_extraction}. By comparing the orientation features of the output motion and the input SOS, the difference can be minimized through iterative optimization using gradient descent.
Specifically, each gradient descent step is given by:
\begin{equation}
\mathbf{p}^* = \mathbf{p}^* - \nabla_{\mathbf{P}} || \mathcal{M}_\mathbf{d}(\mathbf{\hat{q}}) - \mathbf{d}||_2.
\end{equation}
Here, $\mathcal{M}_\mathbf{d}$ denotes the SMS mask of the input SOS script $\mathbf{d}$, ensuring that the orientation feature difference is evaluated only on the visible salient regions.

\begin{figure*}[t]
    \centering
    \includegraphics[width=0.8\linewidth]{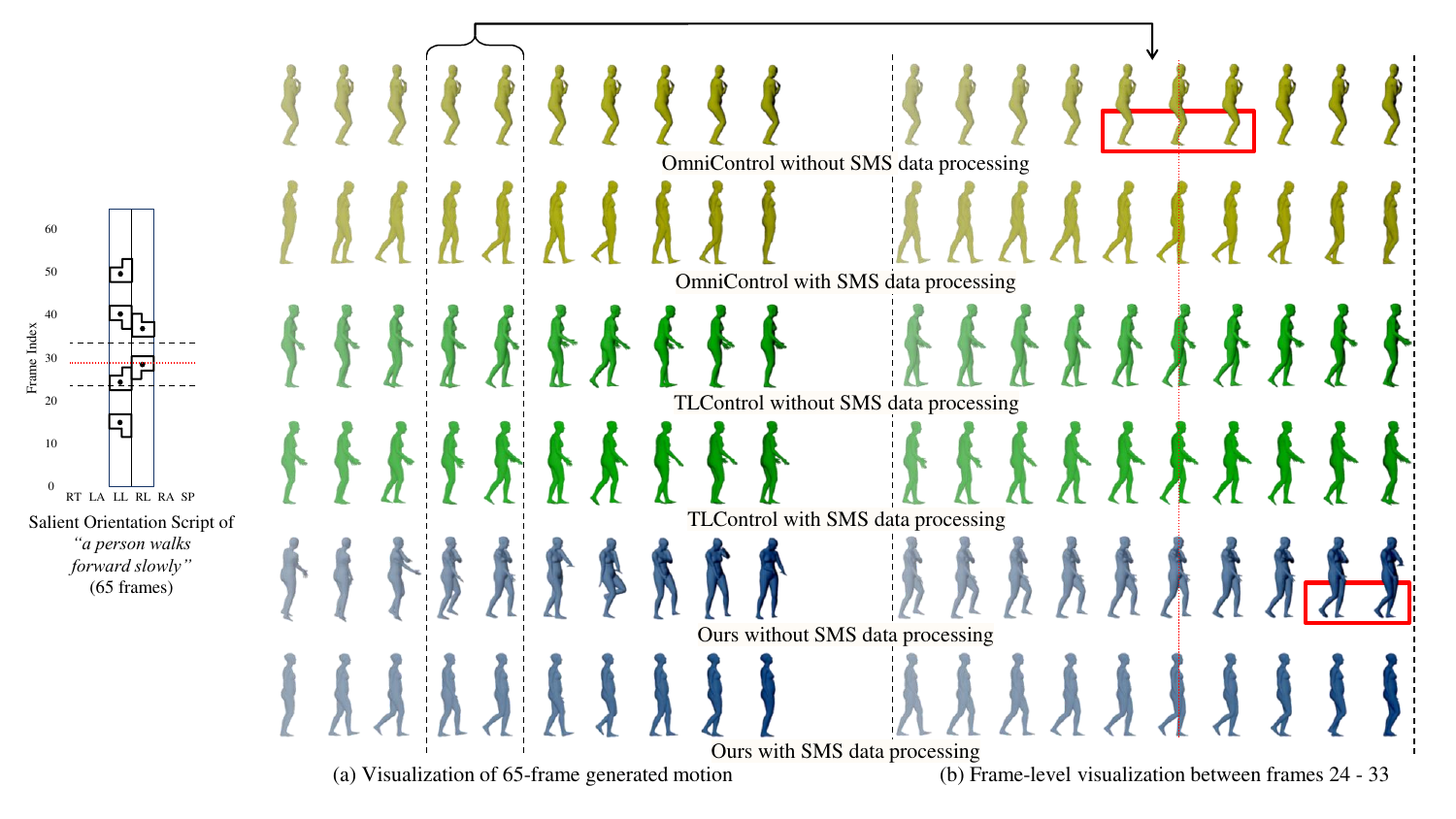}
    \caption{Visualization of SOS-conditioned motion generation results with and without SMS training data processing. (a) presents the motion sequences from left to right, while (b) provides a detailed view of the range between frames 24 and 33.}
    \label{fig:mot_sal}
\end{figure*}

\subsection{Saliency-based Training Data Augmentation}
Following the SOS extraction pipeline in Fig.~\ref{fig:MCS_extraction}, synthesizing the final SOS scripts requires selecting an appropriate saliency threshold for each body part. However, since user-provided SOS scripts can vary in granularity, training with a fixed threshold is suboptimal. To address this, we use SMS-based mask sampling to generate scripts at multiple levels of granularity. For each body part (e.g. \textit{Root} (RT)), we uniformly sample a saliency percentile, $m^{RT} \sim \mathcal{U}(0,1)$, as the threshold for SMS-based SOS script synthesis. This data augmentation across all six body parts exposes the model to diverse saliency variations, enabling it to generalize to a wide range of body part and symbol combinations in user-provided SOS scripts and improving motion generation performance under sparse symbolic control.

\begin{figure*}[t]
    \centering
    \includegraphics[width=0.8\linewidth]{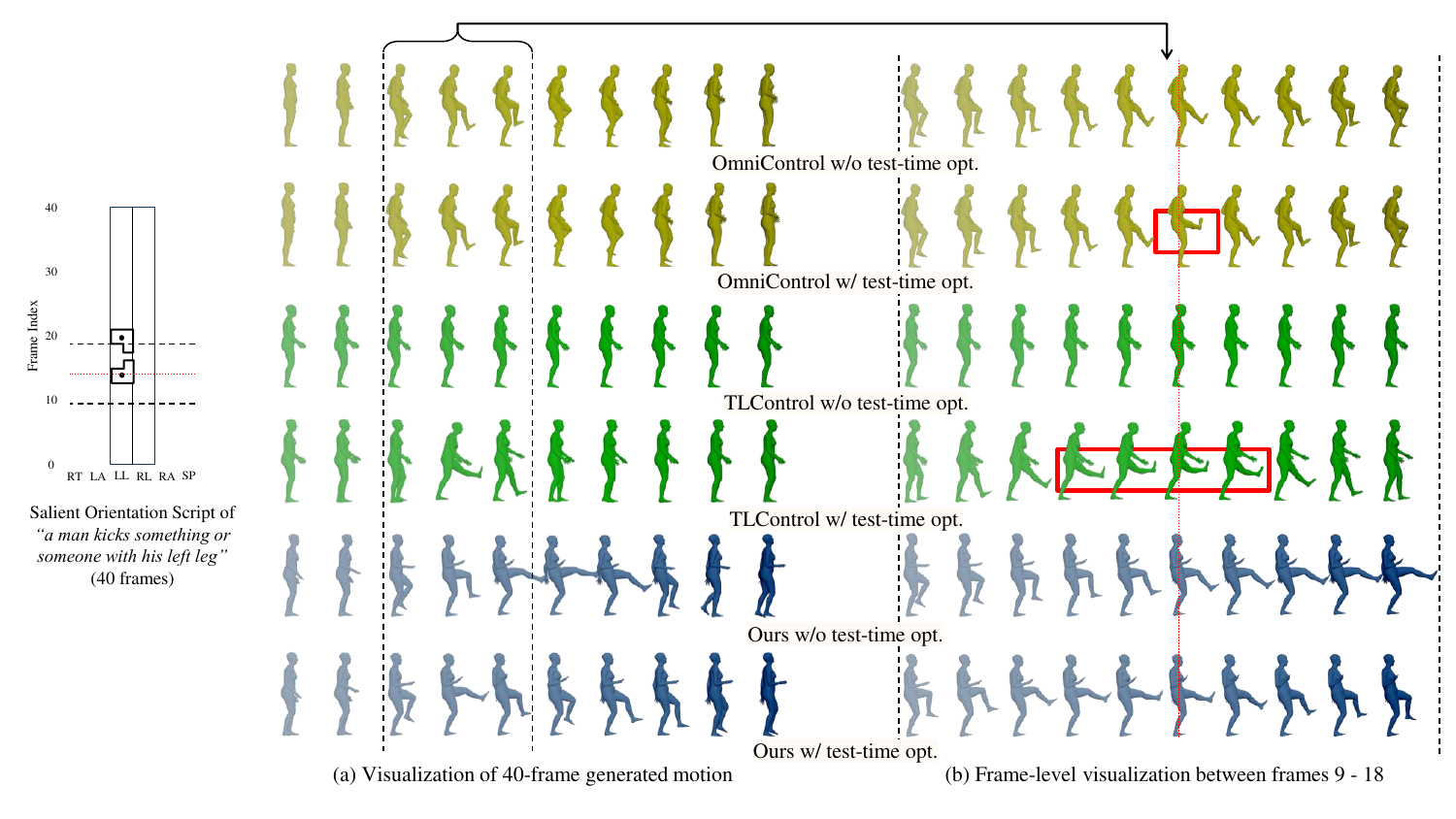}
    \caption{Visualization of SOS-conditioned motion generation results with and without test-time iterative optimization. (a) presents the motion sequences from left to right, while (b) provides a detailed view of the range between frames 9 and 18.}
    \label{fig:iterative_optimization}
\end{figure*}

\section{Experiments}
\subsection{Datasets and Evaluation Metrics}
We utilize the HumanML3D~\cite{guo2022t2m} dataset for training and evaluation. Following the data processing methods from related works~\cite{tevet2023mdm,karunratanakul2023gmd,xie2024omnicontrol}, subsequence lengths are set between 40 and 196 frames.
All models are trained on the same dataset and representation to ensure a fair comparison.

For evaluation, we use \textit{Fréchet Inception Distance} (\textit{FID}) and \textit{Multimodal Distance} (\textit{MMD}) from the T2M~\cite{guo2022t2m} evaluation protocol to assess motion quality and text-motion alignment. To measure control signal alignment, we report SOS accuracy (\textit{SOS-Acc}) by comparing generated motions to the target SOS input. Additionally, we use \textit{L2 loss in 6D rotation}~\cite{zhou2019rot6d} (\textit{L2-Rot6D}) to compare the generated motion with the source motion used to produce the SOS input.

\begin{table}[t] 
\caption{Quantitative results for the \textbf{Motion Generation with SOS Control} on the HumanML3D test set. 
\textbf{Bold} and \underline{underline} indicates the \textbf{best} and the \underline{second-best} result.  
}
\label{tab:mcs_gen}
\begin{center}
\scalebox{0.76}{
\begin{tabular}{lcccc}
\toprule
   & SOS-Acc $\uparrow$ & L2-Rot6D $\downarrow$ & FID $\downarrow$ & MMD $\downarrow$  \\ 
\midrule
Baseline MDM~\shortcite{tevet2023mdm} & 0.151 & 0.351 & \textbf{2.592} & \textbf{6.001} \\
GMD 1-stage~\shortcite{karunratanakul2023gmd} & 0.113 & 0.427 & 25.669 & 7.835 \\
GMD 2-stage~\shortcite{karunratanakul2023gmd} & 0.120 & 0.402 & 21.278 & 7.823 \\
OmniControl~\shortcite{xie2024omnicontrol} & 0.873 & \underline{\textbf{0.325}} & 3.975 & \underline{6.095}  \\
 - w/o SMS data proc. & 0.225 & 0.505 & 60.966 & 8.483 \\
TLControl~\shortcite{wan2024tlcontrol} & 0.982 & 0.341 & 11.132 & 7.066\\
 - w/o SMS data proc. & 0.980 & 0.345 & 13.881 & 7.344 \\
\cmidrule(lr){1-5}
Ours & \underline{0.988} & \underline{\textbf{0.325}} & \underline{3.892} & 6.199  \\
 - w/o SMS data proc. & \textbf{0.991} & 0.499 & 13.494 & 6.893 \\
\bottomrule
\end{tabular}}
\end{center}
\end{table}

\subsection{Comparison Models}
We compare the performance of our method with trajectory-conditioned motion generation models described in the Related Work Section, including GMD, OmniControl, and TLControl, as well as the baseline MDM model, which is conditioned solely on text. Note that GMD 1-stage aligns motion directly to the input SOS script via direct diffusion-time optimization, whereas GMD 2-stage first generates a per-frame quantized orientation script before applying diffusion-time optimization.
To ensure a fair comparison, we adapt these trajectory-conditioned models to utilize the SOS data representation and adjust their ControlNet adaptation and iterative optimization settings as appropriate. Detailed descriptions of the comparison models are provided in the supplementary material.

\subsection{Motion Generation with Saliency Control}
The objective of this experiment is to generate motion conditioned on SOS. In the experimental setup, SOS is extracted from each motion in the test dataset using a saliency threshold of 0.9, and both the SOS and the corresponding textual conditions are provided as inputs to all evaluated methods.

The experimental results, summarized in Tab.\ref{tab:mcs_gen}, demonstrate that our method outperforms others in terms of control signal alignment, as evidenced by superior \textit{SOS-Acc} scores. This highlights our method's effectiveness in leveraging information from SOS, resulting in generated motions that closely resemble the source motions, as indicated by \textit{L2-Rot6D}. In contrast, the baseline MDM obtains the best \textit{FID} and \textit{MMD} scores, as it generates motions unconditionally and is not required to adapt outputs based on SOS. Additionally, Fig.\ref{fig:mot_sal} shows that the generated motions are well-aligned with the body part orientations specified in the input SOS while maintaining natural movement. 

We further assess the impact of saliency in the SOS script during model training. Specifically, we replace the SMS-based training data processing with random masking at a ratio of 0.8 on masking quantized orientation features (w/o SMS data proc.). As shown in Tab.~\ref{tab:mcs_gen}, omitting SMS results in degraded \textit{L2-Rot6D}, \textit{FID}, and \textit{MMD} scores. Additionally, OmniControl, which does not utilize test-time iterative optimization, achieves lower \textit{SOS-Acc} compared to TLControl and our method, both of which employ this optimization. As illustrated in Fig.\ref{fig:mot_sal}, when the SOS script specifies that the \textit{Right Leg} should swing back and reach its peak at frame 29, both OmniControl and our method without SMS data processing fail to achieve the movement peak at the designated frame, treating the orientation symbol as an intermediate waypoint instead. These results highlight the importance of SMS for accurately aligning motion with salient events.

\begin{table}[t] 
\caption{Quantitative results for the impact of diffusion-time and test-time iterative optimizations. \textbf{Bold} and \underline{underline} indicates the \textbf{best} and the \underline{second-best} result.  
}
\label{tab:iterative_optimization}
\begin{center}
\scalebox{0.75}{
\begin{tabular}{lcccc}
\toprule
   & SOS-Acc $\uparrow$ & L2-Rot6D $\downarrow$ & FID $\downarrow$ & MMD $\downarrow$  \\ 
\midrule
GMD 1-stage~\shortcite{karunratanakul2023gmd} \\
 - no opt. & 0.111 & 0.425 & 24.999 & 7.871 \\
 - diff. opt. & 0.113 & 0.427 & 25.669 & 7.835 \\
OmniControl~\shortcite{xie2024omnicontrol} \\
 - no opt. & 0.674 & 0.325 & 4.206 & 6.159 \\
 - diff. opt. & 0.873 & 0.325 & 3.975 & 6.095  \\
 - test. opt. & 0.956 & \underline{\textbf{0.323}} & \textbf{2.782} & \underline{5.992} \\
 - both opt. & 0.956 & \underline{\textbf{0.323}} & \underline{3.025} & \textbf{5.988} \\
TLControl~\shortcite{wan2024tlcontrol} \\
 - no opt. & 0.162 & 0.334 & 42.012 & 8.015 \\
 - test. opt. & 0.982 & 0.341 & 11.132 & 7.066 \\
\cmidrule(lr){1-5}
Ours \\
 - no opt. & 0.531 & 0.329 & 5.570 & 6.382 \\
 - diff. opt. & 0.535 & 0.329 & 5.187 & 6.335 \\
 - test. opt. & \underline{\textbf{0.988}} & 0.324 & 4.209 & 6.125 \\
 - both opt. & \underline{\textbf{0.988}} & 0.325 & 3.892 & 6.199 \\
\bottomrule
\end{tabular}}
\end{center}
\end{table}

\subsection{Ablation study of Iterative Optimization}
We assess the effects of diffusion-time and test-time iterative optimizations on control signal alignment by evaluating four settings for all applicable comparison methods: no iterative optimization (no opt.), diffusion-time optimization (diff. opt.), test-time optimization (test. opt.), and both optimizations applied (both opt.). As shown in Tab.\ref{tab:iterative_optimization}, test-time optimization delivers the most substantial improvements across all metrics. Diffusion-time optimization offers only marginal gains, likely because its adjustments can be overwritten in the diffusion model inference, limiting alignment to the input control signal. Notably, OmniControl achieves the best performance when both optimizations are applied. However, direct test-time optimization on the raw motion signal produces sparse guidance that affects only specific body part keyframes without propagating to adjacent frames, leading to inconsistencies between the modified part and overall motion. Fig.\ref{fig:iterative_optimization} illustrates these effects: test-time optimization in OmniControl only impacts frame 14, causing motion inconsistencies despite favorable metric values. In contrast, our method produces motions that closely follow the SOS specification and maintain overall consistency, benefiting from the ACTOR-PAE Decoder's ability to propagate sparse guidance to neighboring frames. While TLControl’s VQ-VAE decoder can also facilitate propagation, its limited codebook size restricts the expressiveness of the generated motions, resulting in less smooth outputs.

\subsection{Additional Experiments and User Studies}
In addition to experiments on SOS-conditioned motion generation and iterative optimization, we evaluate the effectiveness of each proposed module, analyze model performance under different saliency thresholds, and conduct experiments on the BABEL dataset. We also perform user studies to assess the interpretability of SOS extraction scripts and to evaluate the quality and control alignment of the generated motions. Detailed results from these experiments and user studies are presented in the supplementary material.

\section{Conclusion}
In this paper, we address the limitations of traditional text-conditioned human motion generation by introducing the Salient Orientation Symbolic (SOS) script, a novel and programmable framework for specifying body part orientations and motion timing. By automatically extracting high-saliency keyframes and integrating orientation symbols, our approach enables the generation of saliency-aware sparse symbolic script, providing interpretable visualization and an intuitive, user-programmable symbolic interface for fine-grained motion control. The SOSControl framework further enhances this process by employing saliency-aware data augmentation and gradient-based iterative optimization, ensuring that generated motions closely align with intended orientations and timings while maintaining naturalness and smoothness. By bridging the gap between symbolic motion representation and language-guided generation, our contributions establish a foundation for more efficient, expressive, and accessible workflows in animation, robotics, and interactive human-AI collaboration.

\section{Acknowledgments}
This research was supported by the Theme-based Research Scheme, Research Grants Council of Hong Kong (T45-205/21-N), and the Guangdong and Hong Kong Universities “1+1+1” Joint Research Collaboration Scheme (2025A0505000003).

\bibliography{myRefs}

\newpage

\twocolumn[
\begin{@twocolumnfalse}
\begin{center}\LARGE\bf
    Supplementary Material of SOSControl: Enhancing Human Motion Generation through Saliency-Aware Symbolic Orientation and Timing Control
\end{center}\par
\vspace{1cm}
\end{@twocolumnfalse}
]

\section{Implementation Details}

\subsection{Details of the SOSControl pipeline}
The modules proposed in our framework—including those within ACTOR-PAE and the Motion Latent Diffusion Model—are based on a transformer encoder architecture, configured with: \textit{nhead=8, dim\_feedforward=1024, dropout=0.1}.
All modules are implemented in PyTorch and optimized using the AdamW optimizer with a learning rate of $1 \times 10^{-4}$. The ACTOR‐PAE encoder and decoder are each composed of 8‐layer transformer encoders, with an encoded phase latent channel size of 512. These components are trained for 4000 epochs with a batch size of 512. The denoising transformers in the Motion Latent Diffusion Model also consist of 8 layers and are trained for 4000 epochs with a batch size of 512. For text input, a pre‐trained and frozen \textit{CLIP-ViT-B/32} model is used as the encoder.
All modules are trained on two Nvidia 4090 GPUs, with the training process completed within one day.

After training the ACTOR-PAE and Motion Latent Diffusion Model, we apply ControlNet adaptation following the methodology outlined in OmniControl~\cite{xie2024omnicontrol}. This adaptation extends the transformer encoder architecture by incorporating a series of MLPs to process additional control signal inputs. As detailed in the main paper, both the ACTOR-PAE and Motion Latent Diffusion Model are augmented with this ControlNet adaptation. The extended ControlNet components are subsequently trained for 3000 epochs on two Nvidia 4090 GPUs for one day.

For the diffusion step configuration, we employ DDIM~\cite{song2020DDIM} with 1,000 training steps and 100 inference steps. The number of diffusion-time iterative optimizations is set to 1, and the number of test-time iterative optimization is set to 100. Note that diffusion-time iterative optimization is performed at each diffusion inference step, resulting in a total of 100 iterative optimization executions during inference.

\subsection{Details of Temporal Saliency Detection}
As described in the main paper, we employ temporally‐constrained agglomerative clustering to extract body part frame saliency. The input feature, $\mathbf{o}'$, is derived from the orientation features $\mathbf{o}$ as follows:
\begin{equation}
\mathbf{\hat{o}_t} = \frac{\mathbf{o_t}}{||\mathbf{o_t}||} \cdot \mathbf{u}^T,
\end{equation}

\begin{equation}
\mathbf{o_t'} = \frac{\mathbf{\hat{o}_{t+1}} - \mathbf{\hat{o}_{t-1}}}{2},
\end{equation}
where $\mathbf{\hat{o}}$ represents the normalized dot product between $\mathbf{o}$ and 26 unit‐norm direction vectors, and $\mathbf{o'}$ is computed as the central finite difference of the $\mathbf{\hat{o}}$ at the frame level. The resulting $\mathbf{o'}$ features are then input to the agglomerative clustering algorithm implemented in scikit‐learn, with the following settings: 
\textit{n\_clusters=None, distance\_threshold=0, compute\_full\_tree=True, compute\_distances=True, linkage='ward', connectivity=grid}.

The "grid" in the connectivity parameter refers to a temporal connectivity matrix constructed using the function \textit{grid = grid\_to\_graph(n\_x=T, n\_y=1, n\_z=1)}, where \textit{grid\_to\_graph} is provided by the scikit‐learn feature extraction package. This configuration ensures that cluster merging is restricted to temporally contiguous motion segments, resulting in a temporally connected frame-level hierarchical tree, as illustrated in Fig. 2(c) of the main paper.

\subsection{Hyperparameter}
The primary hyperparameters in our framework are the iterative optimization weights $w_d$ and $w_t$, which scale the gradient term in Equation 6 of the main paper as follows: $\mathbf{p}^* = \mathbf{p}^* - w_d *\nabla_{\mathbf{P}} || \mathcal{M}_\mathbf{d}(\mathbf{\hat{q}}) - \mathbf{d}||_2.$ In our experiments, we set $w_d=300$ and $w_t=3000$.

Other parameters, such as learning rate, batch size, and latent dimension size, were found to have negligible impact on performance during tuning. The specific values for these parameters are provided in the Implementation Details section above.

\subsection{Experiment Details on Comparison Methods}
As discussed in the main paper, our SOSControl pipeline shares a similar model architecture and iterative optimization pipeline with comparison methods. Consequently, similar adaptations of the SOS script can be readily applied to the comparison methods. However, certain methods, such as GMD, require direct imputation of the control signal into the output motion signal. This approach is feasible for global trajectory control tasks, where the output motion representation explicitly contains such features. In contrast, adapting the SOS script for these methods is challenging, since the extracted orientation features do not directly map to any component of the output motion representation.
In this section, we provide details on the adaptations made for each method.

For baseline MDM~\cite{tevet2023mdm}, we follow the standard text-to-motion diffusion framework for model training and motion sequence generation. Note that the SOS script is not incorporated into this baseline pipeline. As a result, the generated motion is not conditioned on the control signal. 

For GMD~\cite{karunratanakul2023gmd}, diffusion‐time optimization is applied with a slight modification from Equation 6 in the main paper. Specifically, the GMD approach computes the gradient term with respect to $\mathbf{p}^k$ rather than $\mathbf{p}^0$ or $\mathbf{p}^*$, which means that the motion latent diﬀusion model is included in the gradient calculation. However, this also restricts the number of diﬀusion‐time iterative optimizations to one. Regarding the imputation process in GMD, it can be applied only during the first stage of the two‐stage pipeline, where per-frame quantized orientation scripts are generated, as direct imputation is feasible due to the direct correspondence between the generated quantized orientation features and the SOS scripts. Conversely, as imputation to the raw motion signal using the SOS script is not possible, we omit the imputation process in both the GMD 1-stage pipeline and the second stage of the GMD 2-stage pipeline during inference.

For OmniControl~\cite{xie2024omnicontrol}, the model architecture and inference pipeline are highly similar to those of our method. In particular, our approach closely resembles OmniControl when ACTOR-PAE is removed, test-time iterative optimization is omitted, and the pipeline operates directly in the raw motion space. Note that the OmniControl pipeline, as described in their paper, does not employ test-time iterative optimization. This design choice is motivated by the sparse guidance issue, which leads to inconsistencies between the modified segments and the overall motion, as discussed in the iterative optimization experiment in the main paper.

For TLControl~\cite{xie2024omnicontrol}, we retain the original model architecture and only modify the control signal input. According to the original TLControl paper, the model conditions on the global three‐dimensional end‐joint positions of five body parts, as well as the root position, when performing global trajectory‐based motion control. This results in a $6 \times 3$ representation, which aligns perfectly with our orientation script design and thus enables direct adaptation.

Finally, for priorMDM~\cite{shafir2024priorMDM}, adaptation to the SOS script is not feasible due to the reliance on both direct imputation of the control signal onto the output motion signal and zeroing out the applied noise in the global trajectory entries. When using the SOS script, the corresponding entries for these operations become ambiguous. Furthermore, removing these steps would reduce the priorMDM pipeline to a form similar to GMD, resulting in redundancy in experimental comparisons. Therefore, priorMDM is not included in our experiments.

To ensure fairness in the diffusion and iterative optimization settings, we set the number of diffusion inference steps to 100, the number of diffusion-time iterative optimization steps to 1, and the number of test-time iterative optimization steps to 100 for all comparison methods, where applicable. 

\subsection{Empirical Running Time of the Pipeline}
In our implementation, the total runtime for generating a batch of motion sequences using our method is approximately 17 seconds. This runtime can be easily reduced by adjusting the number of iterations in the iterative optimization processes. Specifically, motion generation without any iterative optimization takes around 2 seconds. Introducing one diffusion-time iterative optimization step adds approximately 7 seconds, while setting the number of test-time iterative optimization steps to 100 incurs an additional 8 seconds. The primary reason for the increased runtime during iterative optimization is that generating orientation features from motion data requires joint position calculations, which involve executing forward kinematics. This process represents the main bottleneck in the overall runtime. Note that other comparison methods, such as OmniControl and TLControl, exhibit a similar pattern: when diffusion-time and test-time iterative optimizations are applied, the time spent on forward kinematics calculations significantly increases the model inference time.

\begin{table}[t] 
\caption{Ablation studies of the proposed modules on the HumanML3D test set.
\textbf{Bold} and \underline{underline} indicates the \textbf{best} and the \underline{second-best} result.
}
\label{tab:mcs_ablations}
\begin{center}
\scalebox{0.77}{
\begin{tabular}{lcccc}
\toprule
   & SOS-Acc $\uparrow$ & L2-Rot6D $\downarrow$ & FID $\downarrow$ & MMD $\downarrow$\\
\midrule
w/o SMS data proc. & \textbf{0.991} & 0.499 & 13.494 & 6.893\\
w/o ACTOR-PAE & 0.956 & \textbf{0.323} & \textbf{3.025} & \textbf{5.988}\\
w/o ControlNet (Mot.) & 0.986 & 0.326 & 4.126 & 6.133 \\
w/o ControlNet (PAE) & 0.949 & 0.332 & 4.341 & 6.198 \\
w/o ControlNet (both) & 0.956 & 0.333 & 4.611 & 6.258\\
w/o Iter. Opt. (D) & \underline{0.988} & \underline{0.324} & 4.209 & \underline{6.125} \\
w/o Iter. Opt. (R) & 0.535 & 0.329 & 5.187 & 6.335 \\
w/o Iter. Opt. (both) & 0.531 & 0.329 & 5.570 & 6.382\\
\cmidrule(lr){1-5}
Ours & \underline{0.988} & 0.325 & \underline{3.892} & 6.199 \\
\bottomrule
\end{tabular}}
\end{center}
\end{table}

\section{Additional Experiment Results}
This section presents additional experimental results for our proposed method. The results are organized as follows: we first examine the impact of different modules within the SOSControl pipeline. Next, we evaluate the model's performance on motion generation across various saliency thresholds. Furthermore, we assess the influence of varying the hyperparameter associated with the weight of iterative optimization. 
Finally, we conduct experiments using the SOS script on BABEL datasets to investigate how dataset differences affect model performance. 
Except for the experiments involving varying saliency thresholds, all experiments are conducted with the saliency threshold set to 0.9.

\subsection{Ablation Studies on SOSControl Modules}
We evaluate the impact of the proposed modules within the SOSControl pipeline. Specifically, compared to OmniControl, our approach operates in the periodic latent space generated by ACTOR-PAE, incorporates test-time iterative optimization, and integrates ControlNet adaptation into the ACTOR-PAE decoder. As shown in Tab.~\ref{tab:mcs_ablations}, removing ACTOR-PAE results in motion outputs similar to those of OmniControl with test-time optimization. This outcome occurs because removing ACTOR-PAE not only eliminates ControlNet adaptation in the decoder but also causes the model to operate in the raw motion space.
makes the model operate on the raw motion space. Additionally, removing the SMS data processing pipeline or test-time iterative optimization negatively impacts overall performance, as discussed in the main paper. The experimental results further indicate that removing any ControlNet component from the pipeline leads to a slight degradation in motion quality across all evaluated metrics.

\begin{table}[t] 
\caption{Performance of different methods across varying saliency thresholds.
\textbf{Bold} and \underline{underline} indicates the \textbf{best} and the \underline{second-best} result.
}
\label{tab:saliency_ablations}
\begin{center}
\scalebox{0.77}{
\begin{tabular}{lcccc}
\toprule
  Saliency Threshold=0.9 & SOS-Acc $\uparrow$ & L2-Rot6D $\downarrow$ & FID $\downarrow$ & MMD $\downarrow$\\
\midrule
OmniControl~\shortcite{xie2024omnicontrol} & 0.873 & \underline{\textbf{0.325}} & \underline{3.975} & \textbf{6.095}\\
TLControl~\shortcite{wan2024tlcontrol} & \underline{0.982} & 0.341 & 11.132 & 7.066\\
\cmidrule(lr){1-5}
Ours & \textbf{0.988} & \underline{\textbf{0.325}} & \textbf{3.892} & \underline{6.199}\\
\bottomrule
\toprule
  Saliency Threshold=0.7 & SOS-Acc $\uparrow$ & L2-Rot6D $\downarrow$ & FID $\downarrow$ & MMD $\downarrow$\\
\midrule
OmniControl~\shortcite{xie2024omnicontrol} & \textbf{0.990} & \textbf{0.301} & \textbf{1.509} & \textbf{5.743}\\
TLControl~\shortcite{wan2024tlcontrol} & \underline{0.985} & 0.336 & 8.814 & 6.673\\
\cmidrule(lr){1-5}
Ours & 0.983 & \underline{0.306} & \underline{1.870} & \underline{5.851}\\
\bottomrule
\toprule
  Saliency Threshold=0.5 & SOS-Acc $\uparrow$ & L2-Rot6D $\downarrow$ & FID $\downarrow$ & MMD $\downarrow$\\
\midrule
OmniControl~\shortcite{xie2024omnicontrol} & \textbf{0.993} & \textbf{0.276} & \underline{0.804} & \textbf{5.510}\\
TLControl~\shortcite{wan2024tlcontrol} & \underline{0.980} & 0.332 & 8.281 & 6.396\\
\cmidrule(lr){1-5}
Ours & 0.977 & \underline{0.287} & \textbf{0.79} & \underline{5.561}\\
\bottomrule
\toprule
  Saliency Threshold=0.3 & SOS-Acc $\uparrow$ & L2-Rot6D $\downarrow$ & FID $\downarrow$ & MMD $\downarrow$\\
\midrule
OmniControl~\shortcite{xie2024omnicontrol} & \textbf{0.993} & \textbf{0.250} & \underline{0.918} & \textbf{5.379}\\
TLControl~\shortcite{wan2024tlcontrol} & \underline{0.973} & 0.327 & 9.076 & 6.288\\
\cmidrule(lr){1-5}
Ours & 0.947 & \underline{0.265} & \textbf{0.306} & \underline{5.389}\\
\bottomrule
\toprule
  Saliency Threshold=0.1 & SOS-Acc $\uparrow$ & L2-Rot6D $\downarrow$ & FID $\downarrow$ & MMD $\downarrow$\\
\midrule
OmniControl~\shortcite{xie2024omnicontrol} & \textbf{0.982} & \textbf{0.223} & \underline{2.584} & \underline{5.390}\\
TLControl~\shortcite{wan2024tlcontrol} & \underline{0.930} & 0.320 & 8.852 & 6.289\\
\cmidrule(lr){1-5}
Ours & 0.864 & \underline{0.241} & \textbf{0.118} & \textbf{5.318}\\
\bottomrule
\end{tabular}}
\end{center}
\end{table}

\subsection{Ablation Studies on Saliency Threshold Variation}
We also evaluate model performance across different saliency thresholds. Building on the base experiment with a saliency threshold of 0.9, we further assess thresholds of 0.7, 0.5, 0.3, and 0.1. In the base setting (threshold of 0.9), only 17,978 symbols are selected from a total of 3,640,104 dense per-frame quantized orientation symbols across 4,206 test motions, corresponding to a density of 0.0049. For thresholds of 0.7, 0.5, 0.3, and 0.1, the corresponding density values are 0.0106, 0.0189, 0.0386, and 0.1327, respectively. Note that lowering the saliency threshold increases the number of symbols that must be satisfied, thereby making it more challenging to meet the \textit{SOS-acc} condition.   

The experimental results for different saliency thresholds are presented in Table~\ref{tab:saliency_ablations}. Our proposed method demonstrates improved performance in terms of \textit{L2-Rot6D}, \textit{FID}, and \textit{MMD}, although \textit{SOS-acc} decreases as the saliency threshold is lowered. In contrast, OmniControl maintains its \textit{SOS-acc} performance across thresholds, but exhibits poor \textit{FID} at both 0.1 and 0.9 thresholds. This suggests that OmniControl struggles to adapt to SOS scripts that are either too sparse or too dense. 
Finally, TLControl consistently underperforms in both \textit{SOS-acc} and \textit{FID} across different saliency thresholds.

\begin{table}[t] 
\caption{Performance of different methods across iterative optimization weight.
\textbf{Bold} and \underline{underline} indicates the \textbf{best} and the \underline{second-best} result.
}
\label{tab:opt_weight_ablation}
\begin{center}
\scalebox{0.77}{
\begin{tabular}{cccccc}
\toprule
   $w_d$ & $w_t$& SOS-Acc $\uparrow$ & L2-Rot6D $\downarrow$ & FID $\downarrow$ & MMD $\downarrow$\\
\midrule
100 & 3000 & \underline{0.986} & \underline{\textbf{0.325}} & 3.991 & 6.099\\
1000 & 3000 & \underline{0.986} & \underline{\textbf{0.325}} & 4.035 & \underline{6.095}\\
300 & 1000 & 0.958 & 0.326 & 4.246 & 6.156\\
300 & 10000 & 0.952 & 0.332 & \textbf{3.776} & \textbf{6.049}\\
\cmidrule(lr){1-6}
300 & 3000 & \textbf{0.988} & \underline{\textbf{0.325}} & \underline{3.892} & 6.199\\
\bottomrule
\end{tabular}}
\end{center}
\end{table}
\subsection{Experiment on Iterative Optimization Weight}
We also conducted hyperparameter tuning by varying the iterative optimization weights during diffusion-time ($w_d$) and test-time ($w_t$).
The experimental results are shown in Tab.~\ref{tab:opt_weight_ablation}, demonstrating similar motion generation performance across the range of evaluated hyperparameter values. The consistent metrics across different iterative optimization weights indicate stable convergence of the iterative optimization process.

\begin{table}[t]
\caption{Quantitative results for the \textbf{Motion Generation with SOS Control} on the BABEL test set. 
\textbf{Bold} and \underline{underline} indicates the \textbf{best} and the \underline{second-best} result.  
}
\label{tab:babel_gen}
\begin{center}
\scalebox{0.77}{
\begin{tabular}{lcccc}
\toprule
   Saliency Threshold=0.9 & SOS-Acc $\uparrow$ & L2-Rot6D $\downarrow$ & FID $\downarrow$ & MMD $\downarrow$  \\ 
\midrule
Baseline MDM & 0.204 & 0.368 & \underline{0.515} & 3.332 \\
GMD 1-stage & 0.114 & 0.740 & 28.593 & 7.813 \\
GMD 2-stage & 0.109 & 0.722 & 28.780 & 7.705 \\
OmniControl & 0.729 & 0.354 & 0.719 & \underline{3.186}  \\
TLControl & \textbf{0.986} & \textbf{0.340} & 18.892 & 6.673\\
\cmidrule(lr){1-5}
Ours & \underline{0.967} & \underline{0.347} & \textbf{0.205} & \textbf{2.984}  \\
\bottomrule
\end{tabular}}
\end{center}
\end{table}

\subsection{Performance on BABEL dataset}
We also evaluated model performance by training and testing on the BABEL dataset. The primary distinction between HumanML3D and BABEL lies in their annotation schemes. Specifically, BABEL provides annotations and motion segmentations at the atomic action level (e.g., \textit{"raise right arm up"}, \textit{"kick right leg"}), while HumanML3D features natural language descriptions that often combine multiple actions (e.g., \textit{"A person is raising right arm up, then start kicking right leg"}). This difference in annotation granularity generally leads to better \textit{MMD} metric performance for all methods on BABEL compared to HumanML3D. Additionally, the larger number of motion sequences in BABEL contributes to greater variation in \textit{FID} scores across these datasets.

Despite the statistical differences between the datasets, the relative performance of the compared models on BABEL remains consistent with their performance on HumanML3D. For instance, our method achieves competitive results overall, while TLControl consistently outperforms others on the \textit{SOS-Acc} and \textit{L2-Rot6D} metrics, although it performs poorly on \textit{MMD} and \textit{FID}. Conversely, baseline MDM and OmniControl demonstrate strong results on \textit{MMD} and \textit{FID}, but perform suboptimally on \textit{SOS-Acc} and \textit{L2-Rot6D}. These findings are in line with observations from the HumanML3D dataset.

\section{User Studies}
\begin{figure*}
    \centering
    \includegraphics[width=1.0\linewidth]{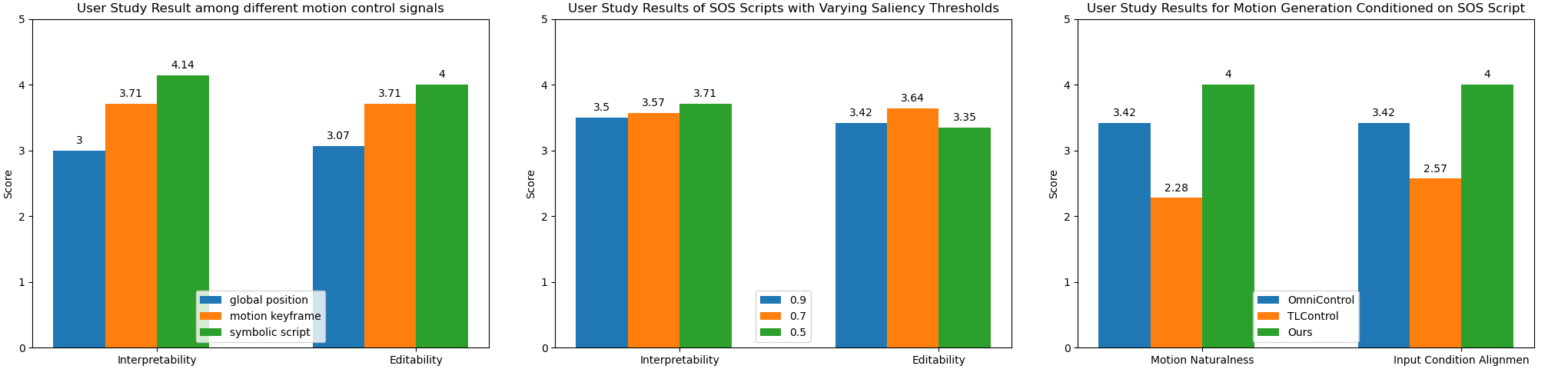}
    \caption{Results from three distinct user evaluations are presented. The left figure illustrates the evaluation of interpretability and editability across different motion control signals. The middle figure displays the results for interpretability and editability of SOS scripts visualized with varying saliency thresholds. The right figure presents the evaluation of motion generation performance for different generation methods conditioned on SOS scripts.}
    \label{fig:user_study}
\end{figure*}

We conducted user studies to evaluate: (1) the effectiveness of the SOS script as a motion control signal for human users; (2) the optimal saliency threshold for enhancing human interpretability and editability of motion; and (3) the degree to which the generated motion aligns with human perception and satisfies the specified input conditions. For each study, we provided visualization figures and motion videos for participants to review and assess. Responses were collected from 15 independent judges, and the results are summarized in Fig.~\ref{fig:user_study}. Further details of the user study methodologies and findings are presented in the following sections.

\subsection{User Study on the Effectiveness of SOS Scripts as Motion Control Signals}

In this user study, we aim to compare our SOS script with global keyframe joint positions and motion keyframes as motion control signals. Five motion clips were randomly selected to generate control signal visualizations and editing videos for the study. To assess control signal interpretability, we presented participants with visualizations of each control signal type and asked them to evaluate the correspondence between the rendered figures and the associated motion sequences. For editability evaluation, professionals were tasked with programming each type of control signal from scratch using relevant tools (e.g., Blender for joint positions and Cascadeur for motion keyframes), with their progress recorded in videos. Participants then reviewed these editing videos and were asked to judge which control signal was easier to program. Representative examples of control signal visualizations and editing videos are included in the supplementary material.

Each participant was presented with 2 out of the 5 available samples, selected at random. Participants were then asked to rate each sample on a scale of 1 to 5 (with 5 indicating the highest quality) across the two aspects described below:
\begin{itemize}
\item \textbf{Interpretability.} On a scale of 1 to 5, where 1 means "unrelated" and 5 means a "perfect match," how would you rate each representation in terms of how well it conveys the motion context?
\item \textbf{Editability.} On a scale of 1 to 5, where 1 indicates "not convenient" and 5 denotes "very convenient," how convenient is each representation for a human user to edit?
\end{itemize}

The user study results are shown on the left of Figure~\ref{fig:user_study}, demonstrating the effectiveness of the SOS script in representing motion for human interpretation, largely due to its ability to depict motion in a two-dimensional symbolic staff. In contrast, the visualization of motion keyframes can be hindered by pose overlapping, especially when the avatar remains in the same position. Programming motion keyframes is also labor-intensive because of the large number of joints that must be controlled, even with the assistance of tools such as Cascadeur. Finally, global keyframe joint positions are difficult to interpret when the motion content of the keyframes is not displayed, as users can only see a few dots on the screen. Moreover, editing these positions requires frequent changes in camera viewpoint, making the programming process cumbersome.

\subsection{User Study of SOS Scripts at Various Saliency Thresholds}

In this user study, 30 motion clips were randomly selected to generate SOS scripts using saliency thresholds of 0.9, 0.7, and 0.5. For each sample, all three symbolic scripts, along with a video of the associated motion sequence, were presented to participants, who were asked to answer two questions. 

Each participant was presented with 10 out of the 30 available samples, selected at random. Participants were then asked to rate each sample on a scale of 1 to 5 (with 5 indicating the highest quality) across the two aspects described below:
\begin{itemize}
\item \textbf{Interpretability.} On a scale from 1 to 5, where 1 represents "unrelated" and 5 represents a "perfect match," how would you rate each symbolic script at different saliency thresholds for its effectiveness in conveying the motion context?
\item \textbf{Editability.} On a scale of 1 to 5, where 1 indicates "not convenient" and 5 denotes "very convenient," how convenient is each representation for a human user to edit?
\end{itemize}

The user study results are shown in the middle of Figure~\ref{fig:user_study}, indicating that all three selected saliency thresholds yield comparable levels of interpretability and editability. These findings suggest that setting a fixed threshold for all motion types may not be optimal. Instead, adapting the saliency threshold according to the motion context could help ensure that the final SOS script is not too verbose, which could potentially confuse human users, while still containing sufficient information to represent the motion context effectively.

\subsection{User Study on the Alignment of Generated Motion with Human Perception and Input Conditions}

In this user study, 30 motion clips were randomly selected to generate motion using OmniControl, TLControl, and our proposed SOSControl pipeline. The input SOS scripts were synthesized with a saliency threshold of 0.9. For each input SOS script, motion videos showing the generated results from all three methods were presented to participants, who were asked to answer two questions.

Each participant was presented with 10 out of the 30 available samples, selected at random. Participants were then asked to rate each sample on a scale of 1 to 5 (with 5 indicating the highest quality) across the two aspects described below:
\begin{itemize}
\item \textbf{Motion Naturalness.} On a scale of 1 to 5, where 1 represents very unnatural and 5 signifies very natural, how would you rate the naturalness and realism of the motions? Are these motions created by a machine or performed by a real human?
\item \textbf{Input Condition Alignment.} On a scale of 1 to 5, where 1 indicates "unrelated" and 5 denotes "a perfect match," how well do the generated motions correspond to the text descriptions and symbolic scripts? 
\end{itemize}

The user study results are shown on the right of Figure~\ref{fig:user_study}, indicating that our model outperforms the others in generating natural motions that accurately align with the input text semantics and SOS script conditions. While OmniControl also produces satisfactory motions, it occasionally exhibits motion artifacts, which negatively impact both motion naturalness and condition alignment. Finally, the motions generated by TLControl appear less natural and less smooth, primarily due to the limited expressiveness of the VQ-VAE encoder.


\section{Limitations and Future Works}
\subsection{Challenge of Adaptive Saliency Thresholds}
One limitation of our approach is the challenge of adaptively determining the saliency threshold according to different motion contexts for effective visualization. Striking a balance between presenting sufficient detail for human comprehension and maintaining enough abstraction to prevent confusion necessitates a nuanced understanding of human cognition. To address this, we plan to incorporate our method into a visualization tool that allows users to interactively adjust the saliency threshold. By utilizing a Reinforcement Learning with Human Feedback (RLHF) framework, we will collect user adjustment data to help identify the optimal saliency threshold for each motion, based on direct human feedback.

\subsection{Challenges in Handling Infeasible Instructions}
Our method exhibits the same failure modes common to all approaches employing test-time optimization, such as TLControl. These methods typically assume that input control signals are physically plausible. However, when the control signals violate kinematic realism—such as requiring both legs to be elevated simultaneously or an arm to transition from a forward to a backward orientation within a single frame—the test-time optimization process enforces adherence to these infeasible instructions, resulting in unrealistic motion generation. To address this limitation, future work will focus on integrating a physical plausibility prior into the test-time optimization process. By learning and encoding constraints from real human motion data, the optimization could automatically detect and propose corrections for kinematically infeasible control signals during inference, thereby ensuring more realistic motion outcomes.

\subsection{Challenges in Generalizing Saliency Detection}
Our SOS script provides a rough sketch for guiding text-to-motion generation, enabling the model to produce motions with satisfactory content (e.g., correct timing for kicking or accurate temporal repetition patterns for running). However, representing control signals solely with quantized orientations limits the granularity of motion control, making it challenging to specify details such as the precise location of a kicking target. Conversely, directly guiding text-to-motion generation using global trajectories and keyframe locations often fails to address such tasks due to issues like trajectory overshooting. While applying our saliency detection algorithm to process global trajectories appears promising, designing effective global trajectory features as the input of the agglomerative clustering algorithm is non-trivial, often resulting in non-representative salient global locations. Future work will focus on adapting the saliency detection algorithm for global trajectory representation.

\section{Discussions}
\subsection{Potential for Managing Gradual Transitions}
Our saliency framework provides precise control over key motion events, which in turn facilitates the synthesis of smooth, gradual transitions. At lower saliency thresholds (e.g., saliency threshold = 0.5), the region between two salient keyframes is considered to exhibit lower saliency, thereby encouraging the formation of gradual and smooth transitions without abrupt changes. Furthermore, the semantic characteristics of these regions can be effectively guided through the input textual conditions. These capabilities showcase the framework’s potential to generate gradual, yet semantically meaningful, motion transitions.

\subsection{Adaptation to User-Defined Scripts}
Regarding the framework’s capacity to accommodate novel or user-defined SOS scripts, we make the following assumption. Constructing SOS scripts with high sparsity (i.e., a saliency threshold of 0.9) is generally straightforward and intuitive, as specifying only a few key symbols rarely leads to significant deviations from ground-truth motions. In contrast, designing plausible SOS scripts with lower sparsity requires a deeper understanding of human motion. Therefore, we assume that users are capable of providing plausible SOS scripts that are reasonably consistent with ground-truth motions, with the level of detail reflective of their expertise.

In accordance with this assumption, most of our experiments are conducted with a saliency threshold of 0.9, where user inputs are expected to closely resemble those extracted from ground-truth motions. To account for varying levels of user expertise, we further evaluate our framework over a range of saliency thresholds, as shown in Tab.~\ref{tab:saliency_ablations}. These experiments demonstrate that our method remains effective as long as user inputs are reasonable and plausible. However, if the provided SOS scripts are not physically plausible, the method will produce unrealistic motion, as the test-time optimization process enforces adherence to such infeasible instructions, as discussed in the Limitations and Future Works section.

\end{document}